\documentclass[runningheads]{llncs}
\usepackage{graphicx}
\usepackage{comment}
\usepackage{amsmath,amssymb} % define this before the line numbering.
\usepackage{color}

\usepackage[width=122mm,left=12mm,paperwidth=146mm,height=193mm,top=12mm,paperheight=217mm]{geometry}

\begin{document}
% \renewcommand\thelinenumber{\color[rgb]{0.2,0.5,0.8}\normalfont\sffamily\scriptsize\arabic{linenumber}\color[rgb]{0,0,0}}
% \renewcommand\makeLineNumber {\hss\thelinenumber\ \hspace{6mm} \rlap{\hskip\textwidth\ \hspace{6.5mm}\thelinenumber}}
% \linenumbers
\pagestyle{headings}
\mainmatter

\title{Fully Embedding Fast Convolutional Networks on Pixel Processor Arrays}
%\title{Weigh in pixels: Fully Embedding Fast Convolutional Networks on Pixel Processor Arrays} % Replace with your title

% CAMERA READY SUBMISSION
% \begin{comment}
\titlerunning{Fully Embedding Fast Convolutional Networks on Pixel Processor Arrays}
% If the paper title is too long for the running head, you can set
% an abbreviated paper title here
%
\author{Laurie Bose\inst{1} \and
Jianing Chen\inst{2 } \and
Stephen J. Carey \inst{2}\and
Piotr Dudek \inst{2}\and 
Walterio Mayol-Cuevas \inst{1}}
\authorrunning{L. Bose et al.}
% First names are abbreviated in the running head.
% If there are more than two authors, 'et al.' is used.
%
\institute{University of Bristol, Bristol, United Kingdom \and
University of Manchester, Manchester, United Kingdom}
% \end{comment}
%******************
\maketitle

\begin{abstract}
We present a novel method of CNN inference for pixel processor array (PPA) vision sensors, designed to take advantage of their massive parallelism and analog compute capabilities.
PPA sensors consist of an array of processing elements (PEs), with each PE capable of light capture, data storage and computation, allowing various computer vision processing to be executed directly upon the sensor device. 
The key idea behind our approach is storing network weights "in-pixel" within the PEs of the PPA sensor itself to allow various computations, such as multiple different image convolutions, to be carried out in parallel. 
Our approach can perform convolutional layers, max pooling, ReLu, and a final fully connected layer entirely upon the PPA sensor, while leaving no untapped computational resources.
This is in contrast to previous works that only use a sensor-level processing to  sequentially compute image convolutions, and must transfer data to an external digital processor to complete the computation.
We demonstrate our approach on the SCAMP-5 vision system, performing inference of a MNIST digit classification network at over 3000 frames per second and over 93\% classification accuracy.
This is the first work demonstrating CNN inference conducted entirely upon the processor array of a PPA vision sensor device, requiring no external processing.

\keywords{Low-level Vision, PPA, CNN, vision sensor, edge computing}
\end{abstract}

\section{Introduction}
\begin{figure}[t]
    \centering
  \includegraphics[width=0.7\linewidth]{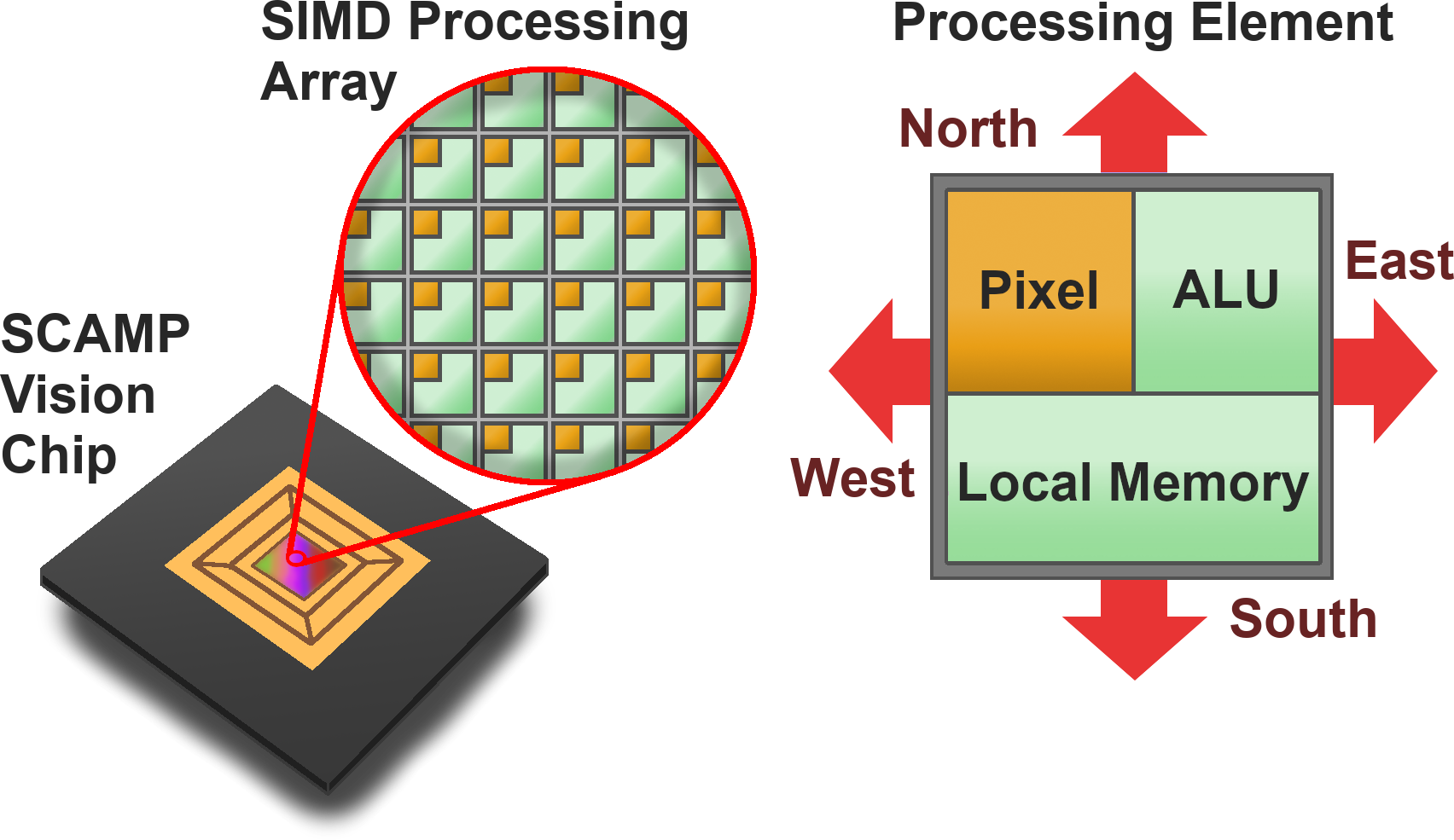}
    \caption{A Pixel Processor Array device performs computations on the image sensor chip, using a SIMD processor array, with each pixel containing arithmetic logic unit (ALU), local memory circuits, and nearest-neighbour communication links.
     \label{fig:scamp5_diagram}}
\end{figure}

Recently, there has been much interest in developing hardware architectures for acceleration of deep learning algorithms. In particular, as Convolutional Neural Networks (CNNs) have become a staple of computer vision applications, there have been many approaches to implementing these efficiently in hardware \cite{du2015shidiannao},\cite{sim2016a142},\cite{aimar2018nullhop},\cite{chen2016eyeriss}. Some of the most challenging application scenarios involve “edge computing” or "on-device computing", where computations are carried out as close to sensors as possible, to achieve low power operation and minimise bandwidth of sensor-processor communications. Ultimately, the sensing and processing can be integrated in a single device. One approach to such integration is through distribution of photosensors of the image sensor within a massively-parallel SIMD cellular processor array \cite{carey2013100},\cite{komuro2004dynamically},\cite{rodriguez2018cmos}, an approach we term Pixel Processor Array (PPA). The PPA concept is illustrated in Figure \ref{fig:scamp5_diagram}.

In areas of computer vision and robotics applications, PPA sensors may potentially offer a wealth of benefits over standard camera sensors that are primarily developed with the human viewer in mind, and designed to capture entire high fidelity images for later inspection. 
The complete image capture, read-out, analog-digital conversion and transfer process in standard sensors introduces a significant time and energy bottleneck in computer vision pipelines, and typically results in low temporal resolution visual information (e.g. typical video-rate of 30 frames per second) that is highly prone to motion blur. 
A PPA sensor circumvents this scenario by instead performing visual computation directly at the point of light capture, extracting the desired information on-sensor, before transferring it over to a host processor.
In many situations this can result in a vast decrease in data bandwidth between the sensor and the external hardware, allowing the system to conduct visual processing at much higher frame-rates, well beyond the capabilities of more standard sensors while, maintaining a low power consumption \cite{carey2013100},\cite{bose2017visual}. %citations 

One application of such PPA sensors is that of neural network inference in which captured visual information is immediately fed through a neural network being executed wholly or partially on-sensor, with the PPA's output then being compressed to simply neuron activations, ideally of the network's final layer.
Such an application of future PPA sensors may offer real world network inference at speeds well beyond standard visual pipelines, however implementation on current PPA hardware is a highly challenging area of investigation.

As an emerging area of research, there exist only a small number of prior works in this area, as discussed in Section \ref{sec:related work}.
These approaches \cite{bose2019camera},\cite{wong2018msc},\cite{guillard2019optimising} suffer from a number of limitations, such as having to perform image convolutions sequentially, requiring certain computation to be performed on external hardware, and only utilizing a small area of the entire processor array.
The work presented in this paper aims to address these issues.
Our main contribution is a new approach for structuring the execution of CNN network inference on PPA architectures.
The key idea behind our approach is the concept of embedding network weights into the "pixels" of the PPA's processor array.
This is done by storing weights within the processing elements (PEs) of the array, rather than weights being contained in the instructions transmitted to the processor array during inference as in previous works.
This embedding of weights allows different parts of the processor array to perform different computations, upon different local data, simultaneously.
As such our approach can perform many different image convolutions, upon multiple images, spread across the PPA array in parallel, and efficiently perform a final fully connected layer entirely on-sensor.
This computation can be structured to make use of the entire processor array at all times, improving the utilisation of available computational resources.
To the best of our knowledge this is the first work to present such an approach, and the first to demonstrate multiple convolutional layers, a fully connected layer, and complete network inference upon a PPA. We demonstrate inference of both 2 and 3 layer networks upon the SCAMP-5 PPA performing digit recognition, able to achieve classifications at over $3000$ frames per second and over 93\% accuracy.

\section{SCAMP-5 Overview \label{sec:scamp}}
\begin{figure}[t]
    \centering
      \includegraphics[width=0.9\linewidth]{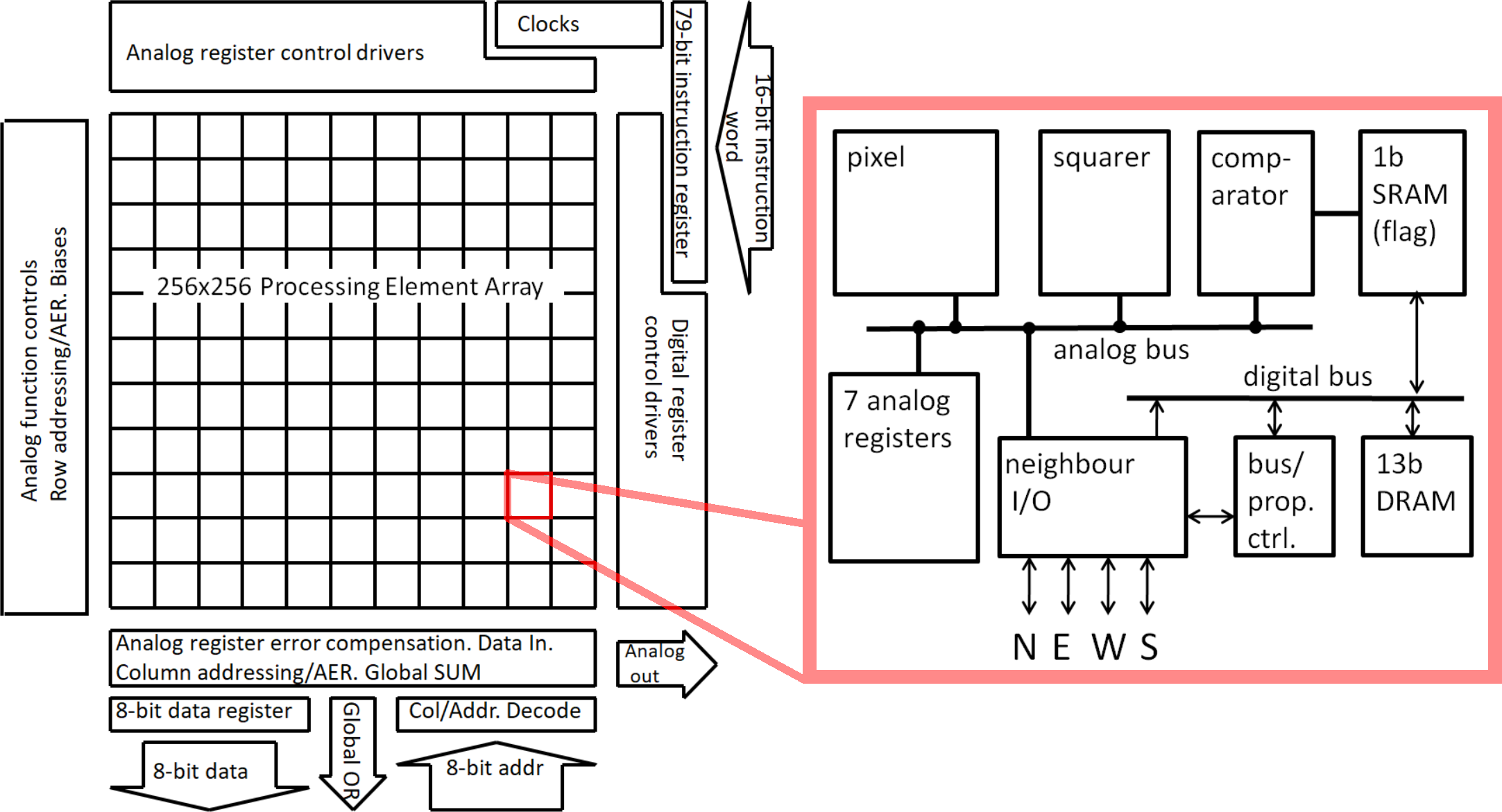}
    \caption{Fig 2. SCAMP-5 Architecture \cite{carey2013100}. Left: The PPA chip contains a 256x256 SIMD processor array and associated control, readout and interface peripheral circuits. Right: The processing element shown contains the photosensor, seven analog local registers, supporting arithmetic operations of addition, negation and division, neighbour communications with 4 nearest neighbour, 1 bit activity flag, and 13 bits of digital memory supporting logic operations. \label{fig:Scamp5 layout}}
\end{figure}

The PPA used for this work is the SCAMP-5 vision sensor \cite{carey2013100},\cite{chen2018scamp5d} consisting of an $256\times256$ array of processing elements (PEs), each containing processor circuitry allowing visual data to be stored, and manipulated directly at the point of light capture. The chip architecture, as shown in Figure \ref{fig:Scamp5 layout}, has been described in  \cite{carey2013100}. Briefly, each PE contains 13 digital registers (1-Bit) and 7 analog memory registers.
% This allows for 13 binary images and 7 gray-scale images of $256\times256$ resolution to be stored upon the sensor.
Various operations can be performed between the memory registers of a PE, such as addition and subtraction of analog registers, and standard Boolean  logic operations between digital registers. PEs can also exchange data with their neighbours. The array operates as an SIMD computer.
The operations on local memory registers are performed across all PEs of the $256\times256$ array in parallel, using a single instruction. Each PE also contains an Execution Flag register allowing it to ignore received operations and allowing for conditional execution. 

The operations performed by the PE array are dictated by a central controller, build upon ARM Cortex M0 processor. This controller executes its own program, primarily for sending instructions to the SCAMP-5 PPA to perform the sequence of operations that will result in some desired computation being performed upon the array.
%Typically, performing a computer vision task upon the PPA involves acquiring images through the photo-sensors on each PE, sending the PE array a series of instructions to process and compute/extract some desired information from this visual data, and finally transmitting this information off-chip to the controller. 

The near-sensor processing approach of this architecture is very efficient. The SCAMP-5 chip performs up to 535
GOPS/W (Giga Operations Per Second per watt). Note that this device is manufactured using two decades old 180nm CMOS silicon technology \cite{carey2013100}. 
Very significant gains can clearly be made on future devices in terms of increasing computing power and decreasing power consumption.

\section{Related Work \label{sec:related work}}
% Here is some text in case we want to add here references to other CNN-hardware, and to some older suff on convolution on sesor.

% Recently, there has been a surge of interest in hardware solutions for near-sensor CNN computation for edge devices. Several dedicated digital processors have been proposed \cite{du2015shidiannao},\cite{sim2016a142},\cite{aimar2018nullhop},\cite{chen2016eyeriss}. In this paper, we focus on methods that integrate the computation into the sensor device itself. Computation of image convolutions with programmable kernels on bespoke sensor devices has been demonstrated before [Gruev, Gottardi, ACE-16k]. More recent PPA devices enable programmable in-pixel computation, enabling a fuller implementation of CNN inference networks.
While previous works exist regarding CNN inference on PPAs \cite{bose2019camera},\cite{wong2018msc},\cite{guillard2019optimising} typically these methods perform various parts of the network computation in serial, rely on external hardware for additional computation, and only make use of a small area of the PPA's processor array leaving a great amount of processing power untapped. 
For example, these approaches are demonstrated upon MNIST/digit classification task on SCAMP-5 in which they load a single small MNIST digit ($28\times28$) into the center of the the $256\times256$ SIMD processor array.
These approaches then sequentially compute image convolutions upon this central digit, effectively leaving well over $90\%$ of the processor array unused. 
These convolution results are passed to the ARM controller connected to the SCAMP-5 PPA, which is used to perform one or more fully connected layers.
Therefore, a significant portion of the neural network computation in these approaches is actually conducted upon the ARM controller in a standard C++ program rather than by making use of the PPA's processing power.
 
By comparison, the approach  proposed in this paper performs complete inference computation, including the fully-connected layer, upon the PPA device, 
%, with no further external computation required. 
% Only the activation values of the final layer of neurons are output from the PPA chip.
%Our approach can 
potentially utilizing 100\% of the processing array, and efficiently performing convolutional layers by computing many different image convolutions in parallel.
%, and is the first work to perform a fully connected layer upon the processing array rather than on external processing.
%This is significantly more efficient than existing approaches, and also does not require additional processing hardware to complete network inference. 

The proposed approach requires all network weights to be stored upon the processing elements (PEs) of the PPA itself. 
Due to the limited memory (13 Bits, 7 analog values) of each PE on current generation PPA hardware, we are restricted to low-bit quantised weights and a limited number of layers.
However it should be noted that many tasks have been successfully demonstrated on such low-bit weight networks \cite{zhu2016trained}\cite{hubara2017quantized}\cite{zhou2017incremental}, and it is likely that next generation PPA hardware will see a significant boost in memory per PE.

% \section{Architecture Overview}

\section{Parallel Convolutional Layer Computation \label{sec:convol layer}}

In this section we describe our approach for the computation of convolutional layers upon the PPA.
The weights of the various convolutional filters are stored upon the processing array, within the registers of the PEs.
This enables different convolutional filters to be applied to different areas of the PE array in parallel.
This can allow us to perform all the computation required for a convolutional layer simultaneously.
%rather than computing convolutions sequentially one after another by executing a sequence of SIMD instructions sent from the controller.

\begin{figure}[t]
    \centering
      \includegraphics[width=0.9\linewidth]{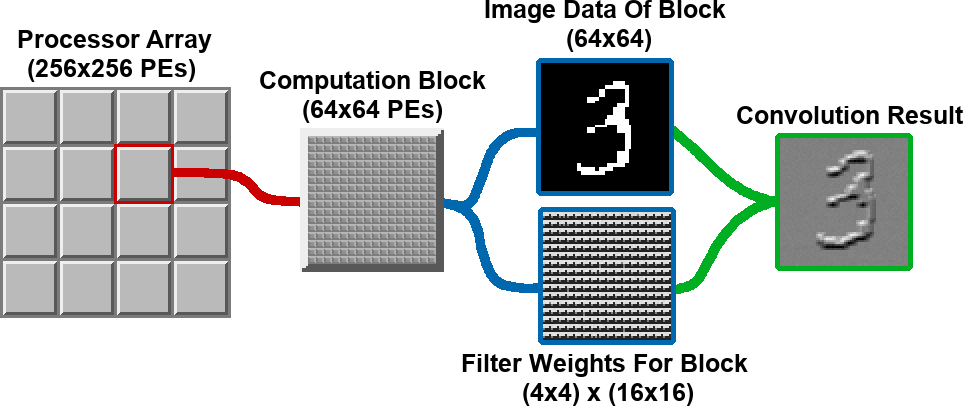}
    \caption{Computational layout of the processing array for a convolutional layer. The array is split into computation blocks, each containing a set of filter weights (duplicated many times within a block) and the image to which the filter will be applied. A single SIMD routine can then be executed to apply each block's filter to its image data in parallel.  \label{fig:convol computation layout}}
\end{figure}

\begin{figure*}[t]
    %TODO make this 8x8 digit grid  
    \centering
        \includegraphics[width=0.99\linewidth]{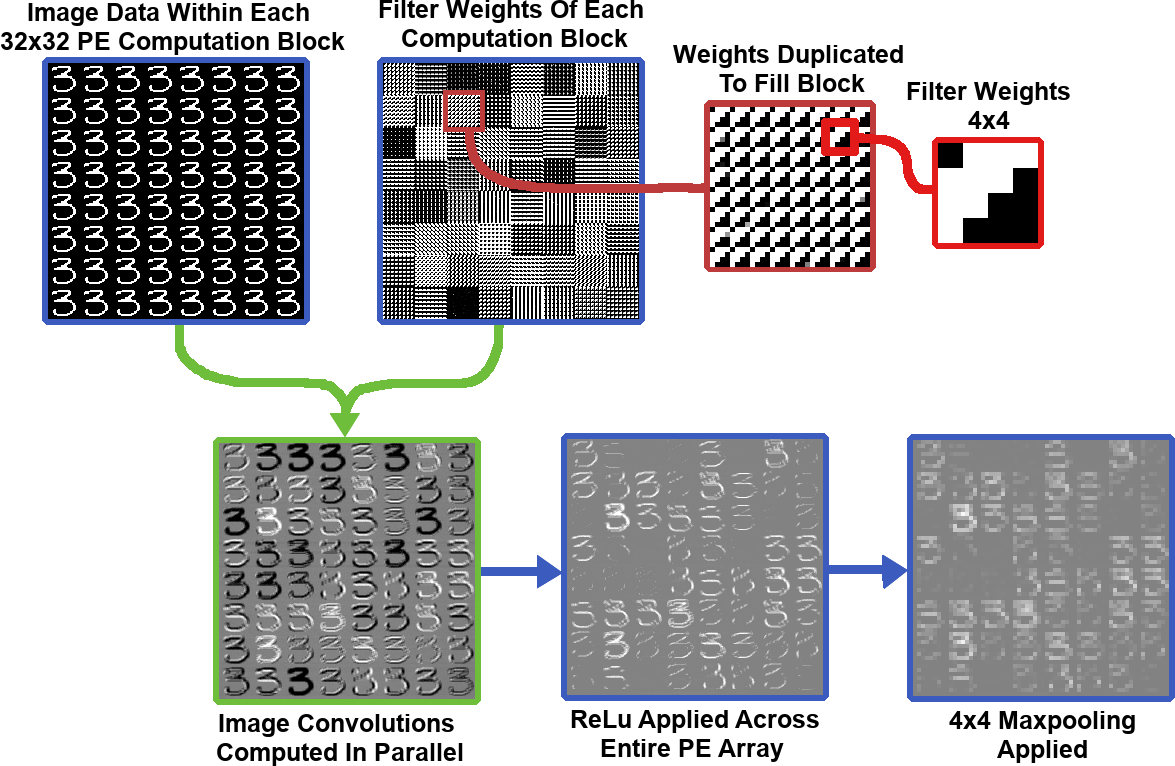}
    \caption{Examples of Convolutional layer computation, illustrating how multiple convolutions are computed in parallel by splitting the PPA array into distinct computation blocks. In this case the Scamp5's processing array is split into 64 computational blocks of $32\times32$ PEs each.
    Upon computing image convolutions ReLu and Maxpooling can also be applied across the array at very little computational cost.
    % Each block stores its own convolution filter weights and image data. An SIMD routine can then be executed to apply each block's filter to it image data in parallel, apply ReLu, and finally max-pooling to produce the layer's activation data.
    \label{fig:convol examples}}
\end{figure*}

For example, in the case of SCAMP-5, up to 64 MNIST digits can be spread across the $256\times256$ PE array. 
This allows for 64 different convolutions to be performed simultaneously at no additional time or power cost.
In the case of digit classification this can be used to compute 64 different convolutions on the same digit duplicated 64 times in parallel.
%something that would be significantly slower in previous works that compute convolutions sequentially \cite{bose2019camera},\cite{wong2018msc},\cite{guillard2019optimising}.

\subsection{Computational Layout On PE Array}

Our convolutional layer approach effectively divides the PE array into multiple rectangular "computation" blocks of processing elements.
The PEs of each computation block contain both the weights of a specific convolutional filter and image data to which the filter should be applied as shown in Figure \ref{fig:convol computation layout}. A sequence of SIMD operations can then be formulated to simultaneously apply each computational block's filter to its stored image data, performing all the computation required for a convolutional layer.
Examples of such computation are illustrated in Figure \ref{fig:convol examples} for MNIST digits.
% The PPA's controller can then simply send instructions for this sequence of operations to perform all the computation required for an entire convolutional layer. 

Note this approach is flexible in that each computational block may contain different image data, and the size and dimensions of each block may also vary, however, for convolutional layer computation we use square blocks of identical size.
%However, in most cases one wishes to compute multiple different convolutions upon the same input data as in the digit recognition task demonstrated by this work.
For digit recognition we demonstrate convolutional layers of both $64$ and $16$ convolutions, using computational blocks of size $32$ and $64$ respectively.
In both cases the $28\times28$ MNIST digits are rescaled to fill these computation blocks.
% In our experiments,  we divide the $256\times256$ array into $64$ equally sized blocks, each of $32\times32$ PEs, and containing a duplicate of the MNIST digit (padded up to $32\times32$ image size) to be classified.

\subsection{In-Pixel Filter Weights}

Each computation block stores within it the weights of a specific filter.
When the SIMD routine for a convolutional layer is sent to the processor array, each block will use these weights to compute a convolution upon its stored image data.
Directly storing filter weights upon the processor array at the locations where they are to be applied is what allows our approach to perform multiple filters simultaneously.

There are many possible layouts for storing a set of filter weights within a computational block of PEs. However, it is generally not possible for each PE to store a complete copy of its block's filter weights due to the limited local memory resources available on current generation PPA devices.
The solution is to spread the storage of a computational block's filter weights across multiple PEs. 
This means each PE no longer has immediate access to every filter weight, however, weights can be copied over from other nearby PEs of the same computational block during convolution computation.
To minimize the time transferring filter weights between PEs, is important to use a layout in which each PE is located in close proximity to other PEs storing the weights it will require during computation.
This prompted a "checker board" style layout, where multiple copies of convolutional filter weights are stored within each computational block to ensure each PE is located within a reasonable distance from each filter weight.
This concept is illustrated for $4\times4$ filters in the right of Figure  \ref{fig:convol examples}.
% Note that this process of having to copy filter weights to all PEs in which they are required is by far the most significant bottleneck during convolution computation.
Future PPA devices, with greater resources per PE, should allow each PE to store its own dedicated copy of any filter weights, significantly speeding up the convolution computation.

In our demonstrated networks each PE in a block stores a single binary filter weight, with the weight values of $+1$ and $0$ naturally corresponding to image addition and subtraction operations.
There are many schemes that could be used to store and apply higher bit-count weights but for now we leave this to future work.

%NOTE THE MOST EFFICENT WEIGHT STORAGE LAYOUT (THAT MINIMIZES TIME OF WEIGHT PROPAGTION TO ALL PE IN A BLOCK) WILL DEPEND ON THE FILTER SIZE AND IMAGE SIZE, THE CURRENT SCHEME IS CERTAINLY NOT THE BEST. THIS IS PROBABLY THE MOST IMPORTANT AREA TO OPTIMIZE 
%----------------ASK JIANING TO ADD "scamp5_kernel_begin(X);" WHICH SIMPLY MAKES ONE KERNEL DUPLICATING THE CODE X TIMES, DOING THIS MANUALLY MAKES THE CODE LOOK LIKE SHIT AND MAKES IT PRONE TO BUGS

% \begin{figure}[t]
%     \centering
%       \includegraphics[width=0.31\linewidth]{Pictures/filter_weights_in_pixel2.jpg}
%     \caption{Format for storing convolutional layer filter weights upon the PE array, with each computation block containing the weights of a different filter. \label{fig:inpixel filter weights}}
% \end{figure}  

\subsection{Parallel ReLU and Max Pooling \label{sec:maxpool relu}}
After performing a convolutional layer the PE array will hold multiple convolution images such as those shown in Figure \ref{fig:convol examples}.
We then turn these images into activation data by first applying the ReLU activation function.
The SCAMP-5 hardware has the function to flag all PEs whose stored values in a certain analog register are positive or negative.
This allows us to simply flag all PEs whose convolution result is negative and input a value of zero into these flagged registers, generating ReLu activations.

% Applying ReLU simply consists of selecting all PEs whose stored convolution result (a single pixel value) is negative, and then replacing these values with 0.
% This is performed in parallel upon the entire array using a small set of SIMD instructions, generating an image of activation values.

We then perform a $4\times4$ max pooling routine by first making a copy of the activation values image.
This copied image is then shifted horizontally right, with each PE then containing both its original activation value and a value from this shifted data.
In parallel, every PE then compares these two activation values, replacing the stored activation data with the shifted data whenever it is greater in value.
This routine of shifting horizontally, comparing activations and replacing with the higher value is repeated three times, followed by a similar routine three times shifting vertically down.
This results in every PE holding the highest activation value in the $4\times4$ square of which it is in the top left corner. 
The pixels holding the correct max-pooled values for each $4\times4$ grid space are then copied back into each PE of their $4\times4$ block. %TODO BETTER EXPLAIN THIS ADD DIAGRAM?

\begin{figure*}[t]
    \centering

   \includegraphics[width=0.99\linewidth]{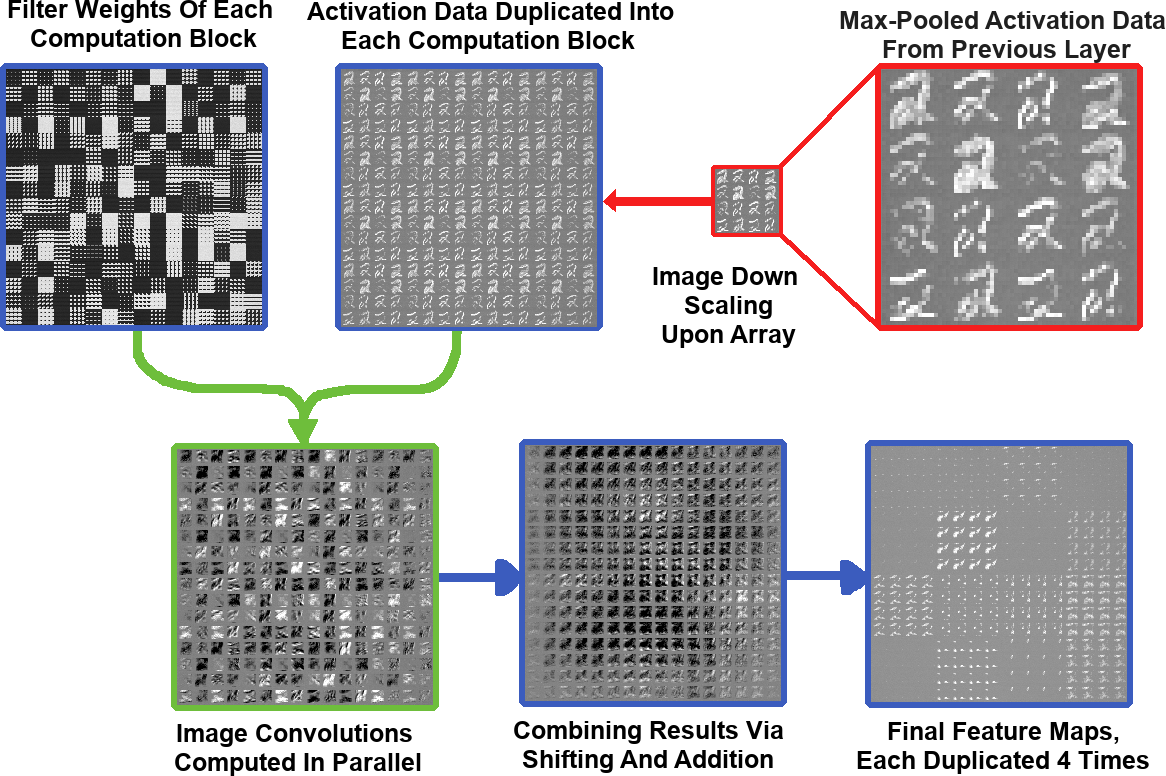}
    \caption{Computation of feature maps from an additional convolutional Layer. 
    Max-pooled activation data from the previous layer (top right) is shrunk and duplicated across the processor array. Image convolutions for the new layer are them performed upon each duplicated image, in the same manner as network's initial convolution layer.
    The resulting convolutional images (bottom left) are then combined accordingly and ReLu is applied forming feature maps of this new layer (bottom right).
    In this example 16 feature maps from the initial layer are duplicated and 256 convolutions computed, before being recombined to form 16 feature maps.
    \label{fig:additional convolution layer diagram}}
\end{figure*}

\subsection{Further Convolutional Layers}

After performing an initial convolutional layer, either a final fully connected layer, or an additional convolutional layer can be performed.
This section describes one possible method to compute such an additional convolution layer, where each feature map is constructed from those of the previous layer as standard.
Note that this approach could in future be used to add multiple additional convolutional layers, however this is difficult to achieve within the limited memory resources of current SCAMP-5 hardware.

In brief, the feature maps of a previous convolution layer (consisting of max-pooled activation data) are shrunk and duplicated to fill the processor array.
Each duplicate of a feature map is then used in computing a feature map in the new convolutional layer.
An example of this is shown in Figure \ref{fig:additional convolution layer diagram}, where 256 convolutions are computed in parallel upon the 16 feature maps (each duplicated 16 times) of the previous layer.
These convolution results are then added together accordingly to form the 16 feature maps of the new convolutional layer.

Many of the concepts introduced previously are re-used for this computation.
The in-pixel storage of filter weights and computational layout is identical to the initial convolutional layer, with the processor array again being split into computation blocks each storing its own set of weights and image data as shown in Figure \ref{fig:additional convolution layer diagram}.
The same SIMD routine used to perform the initial convolutional layer can simply be executed again for computing this additional layer, helping to reduce program size.

The resulting convolution results are then repeatedly shifted and added together, iteratively accumulating feature maps of this new convolutional layer.
These feature maps can then be duplicated across the array, correctly positioning their activation data to be aligned with the weights of any following fully connected layer, so that parallel multiplication between activations and fully connected weights can be performed as described in Section \ref{sec:fc layer}.

\subsection{Feature Map Shrinking and Duplication}

% TODO mention scaling of data to avoid saturation???
The process of shrinking the max-pooled activation data upon the PPA leverages the image transformation methods first introduced in \cite{bose2017visual} for image scaling.
However, conducting such scaling operations using analog memory registers results in the build-up of systematic errors and noise \cite{carey2013100}, from analog data having to be repeatedly copied from one PE to the next.
This would corrupt the activation data beyond use.
To avoid this issue we instead convert the analog activation data to a 3-bit digital representation, with each PE's stored analog value being split across 3 digital registers (within the same PE).
This creates 3 binary images, one for each bit, which then can then all be scaled and duplicated across the array.
Afterwards this digital data can be recombined to once again form a single gray-scale analog image, but devoid of corruption.
%This process is illustrated in figure TODO along with a comparison of performing the same transformation purely in analog without the intermediate digital conversion to demonstrate its necessity.

\begin{figure}[t]
    \centering
\includegraphics[width=0.99\linewidth]{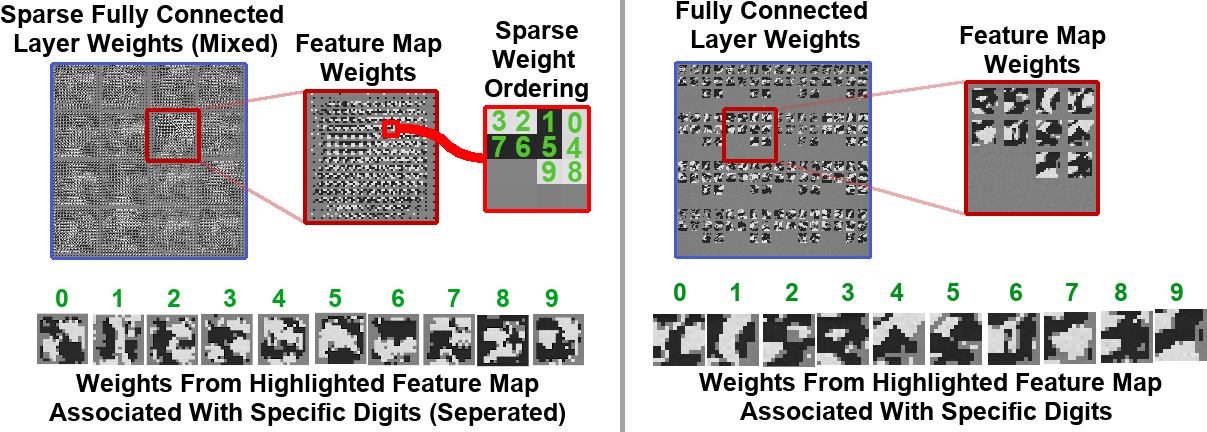}
    \caption{Two examples of ternary fully connected weights stored upon the PPA's processor array in analog memory, both connected to 16 features maps from a previous convolution layer. Left: weights for connecting to ($4 \times 4$) max-pooled activation data stored in a sparse checkerboard like layout, with weights for the different digits mixed/inter-weaved with one another. Right: weights for connecting to duplicated feature maps. \label{fig:fc weights per digit}}
\end{figure} 

\section{Parallel Fully Connected Layer Computation \label{sec:fc layer}}
\begin{figure*}[t]
    \centering
      \includegraphics[width=0.99\linewidth]{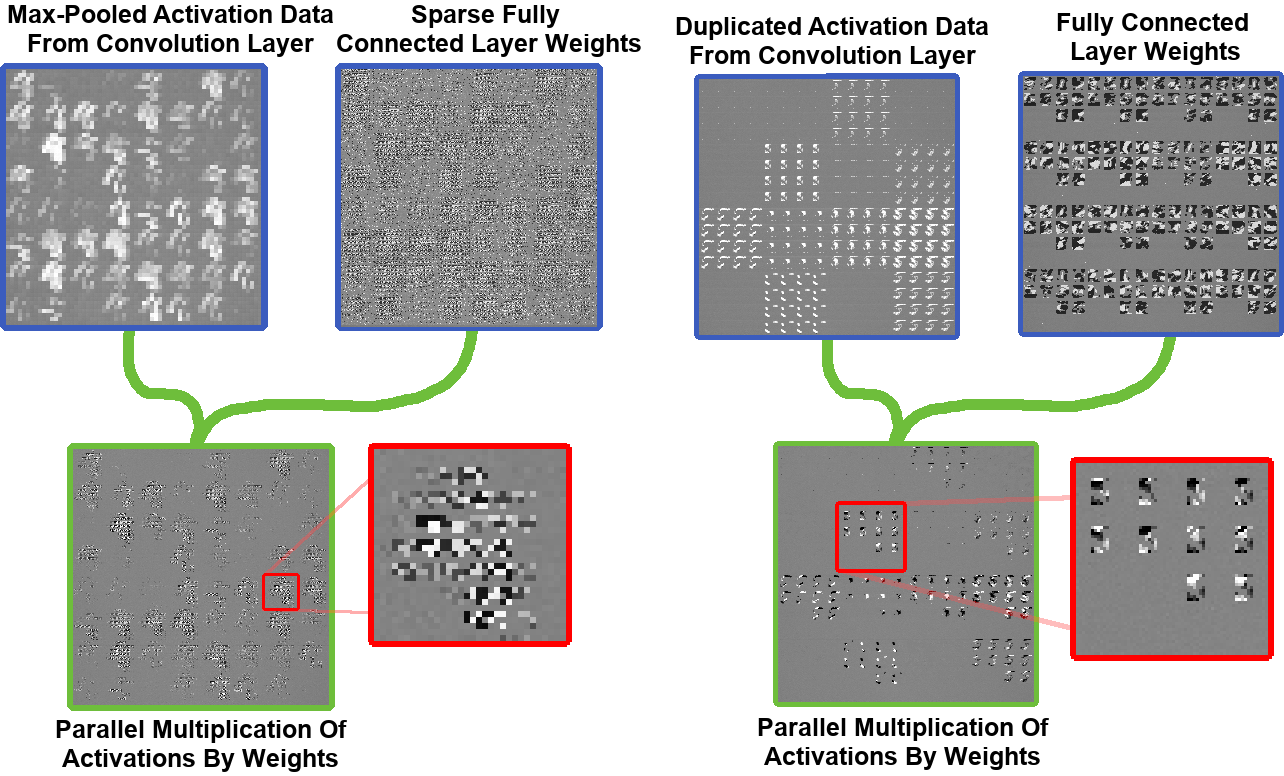}
    \caption{Layout and application of fully connected layer weights, accepting as input either max-pooled activation data (Left) or duplicated feature maps (right). The arrangement of weights is different in each case to enable the correct parallel in-line multiplication between the weights and activation data. After this multiplication, all resulting data associated with a specific fully connected neuron can be summed in parallel by flagging the appropriate PEs.   \label{fig:fc layout diagram}}
\end{figure*}

Following on from a convolutional layers, we perform computation of a final ternary weight fully connected layer upon the PPA, again storing weights directly in the PEs of the processor array.
The activations of the previous convolution layer are duplicated as shown in Figure \ref{fig:fc layout diagram}, either by max-pooling (which creates blocks of duplicated values) or by duplicating the feature maps multiple times across the array.
By correctly arranging the layout of the fully connected weights, each weight's PE can then directly receive the activation data associated with that weight.
This layout varies as illustrated in Figure \ref{fig:fc layout diagram} depending on whether the previous layer produces max-pooled data or duplicated feature maps.
All duplicated activations from the previous layer can then be multiplied by their associated fully connected weights simultaneously in parallel, using the native analog image addition (for weights of value 1) and subtraction (weights of value -1) operations of SCAMP-5.
Examples of this process are shown in Figure \ref{fig:fc layout diagram}.

% \begin{figure*}[t]
%     \centering
%       \includegraphics[width=0.99\linewidth]{Pictures/register_content_FC_layer2.png}
% %   \includegraphics[width=0.31\linewidth]{Pictures/maxpooled_data_contrast2.jpg}
%     \caption{An example of generating synaptic contributions for the final fully connected layer across 64 computational blocks in parallel. From left to right, convolution image data, ReLU and max pooling operations being applied, fully connected weights stored upon the array, fully connected weights multiplied by max-pooled data to produce synaptic contributions.  \label{fig:maxpool example}}
% \end{figure*}  

The limited memory of the processing elements on SCAMP-5 restricts to the use of ternary weights for the final fully connected layer, stored within the analog registers of the PEs to save digital resources. 
Note that the content of analog registers decays over time, drifting away from the stored value \cite{carey2013100}. 
However, with quantized ternary values being stored one can "Refresh" the register's content at set intervals to prevent such decay.

% Figure \ref{fig:fc weights per digit} shows the fully connected weights for digits 0 and 2 specifically, illustrating what structure each of the digit's weights is looking for in the various convolution results from the previous layer. 

\subsection{Activation Value Summation}

After the multiplication step each PE contains a synaptic contribution to the activation of one of the final neurons in the fully connected layer. 
 SCAMP-5 has the capability of performing a global summation of many analogue values distributed across the PE array in parallel, which
 can be used to effectively add all synaptic contributions (thousands of values in this case) for one output neuron in a single clock cycle. 
 This can be used to provide an approximate summation of the activation data for a specific neuron, however, the analog method of summation introduces significant noise.
 
 For two layer networks in which a large number of activations get through to the fully connected layer, this does not pose a major issue. For these networks analog summation can be simply be performed a number times and averaged to aid in noise mitigation.
 However for three layer networks, where features after the second layer have become more discriminative, the analog noise becomes a factor limiting the network performance.
 
\begin{figure*}[t]
    \centering
      \includegraphics[width=0.99\linewidth]{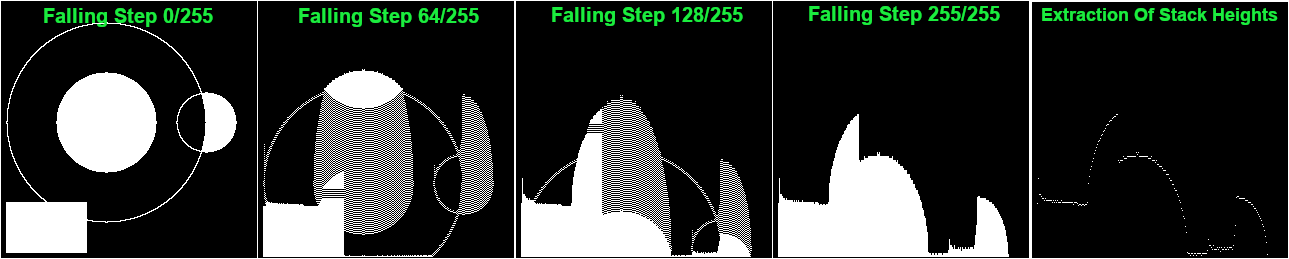}
    \caption{Example of our proposed method for rapid binary image summation being iteratively performed upon a test image. With each iteration the "1s" of the image are made to fall forming stacks at the bottom of the image.  All but the tops of these stacks are then eliminated allowing the heights of each to be read directly.  \label{fig:dsum fall}}
\end{figure*}

 In order to get accurate activation values for such three layer networks we instead turned to using the PPA's digital computation, creating a new method to rapidly count the set white pixels ("1s") in a binary image.
 This method, as visualised in Figure \ref{fig:dsum fall}, functions by essentially stacking pixels together on one side of the array.
 This process can be efficiently implemented upon the PPA's parallel architecture, performing 255 iterations of simultaneously shifting and stacking pixels.
A simple shift copy and XOR can be used to eliminate all but the top pixels of each of these stacks.
The image coordinates of these remaining pixels (up to 256) can then be read directly (using an address-event scheme) to give the heights of each stack, which when added together give the total number of white pixels in the original image.
This entire process takes 260$\mu s$ to complete, and while slower than the analog global summation, it provides a perfectly accurate summation result.
This method is employed in the fully connected layer of our demonstrated three layer network, converting the analog activations into multi-bit representations, which can then be summed.

\section{Results \label{sec:results}}

\subsection{MNIST Network Training \label{sec:training}}

We trained networks with a mixture of binary and ternary weights (for convolutional filters and fully connected weights respectively) using an approach similar as \cite{courbariaux2015binaryconnect},\cite{bose2019camera}, whereby real-valued weights are stochastically quantized during every forward pass.
The errors obtained from the forward pass are then used to update the real-valued weights using the standard error back propagation algorithm, resulting in these real values converging towards binary/ternary ones over the course of training.
%Figure \ref{fig:convol examples} left shows one example of a set of $64$ ternary convolutional filters trained using this approach.

\begin{figure*}[t]
     \centering
     
        \includegraphics[width=0.96\linewidth]{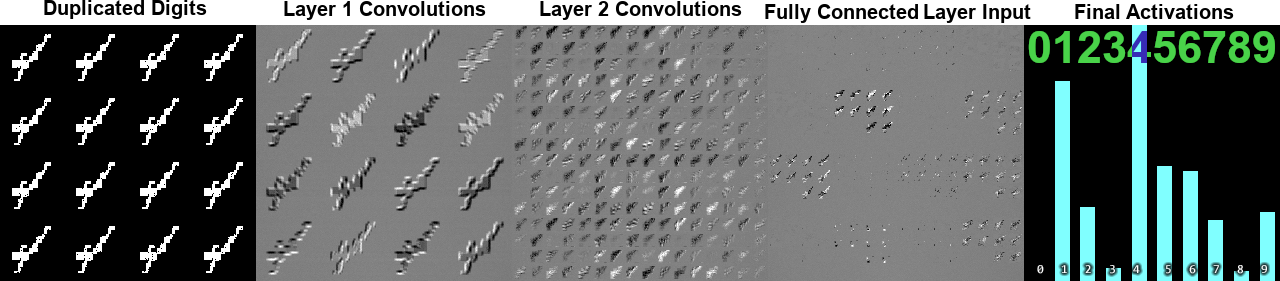}
          \includegraphics[width=0.96\linewidth]{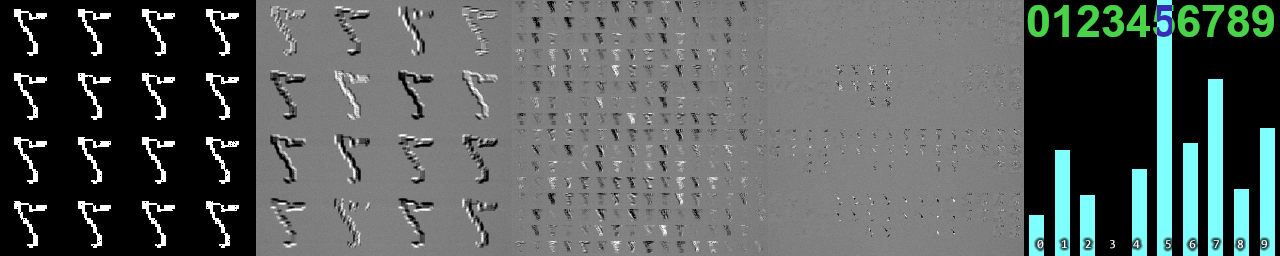}   
           \includegraphics[width=0.96\linewidth]{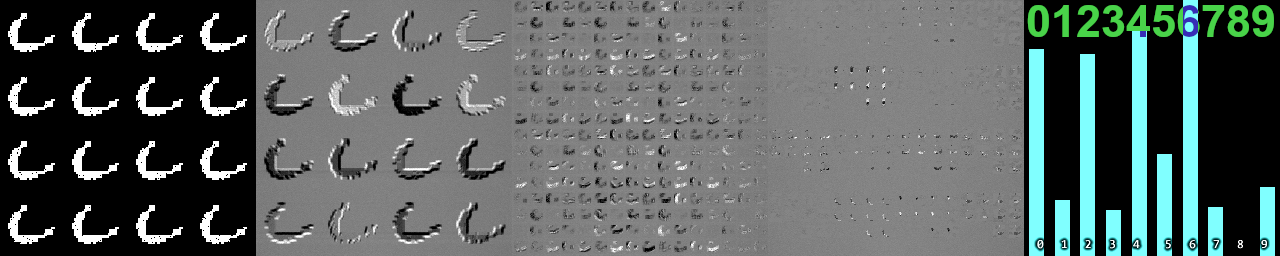}

    \caption{Example classifications of some ambiguous digits via inference of a three layer network upon SCAMP-5. The convolutions computed from both layers are shown, along with the activations multiplied by the fully connected layer weights, and the final neuron activations for digits 0-9. \label{fig:3layer classification examples}}
\end{figure*}

\begin{figure*}[t]
     \centering
      \includegraphics[width=0.96\linewidth]{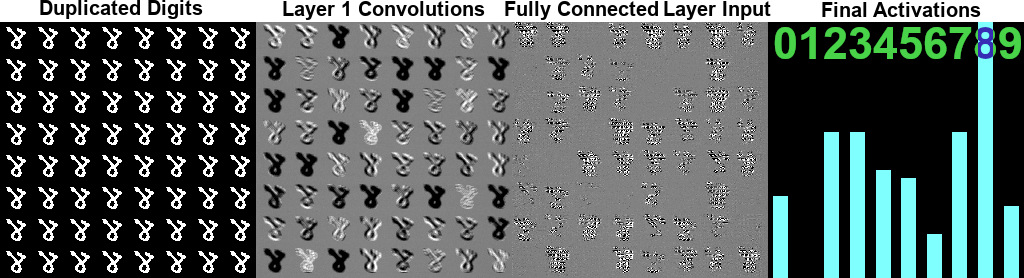}
          \includegraphics[width=0.96\linewidth]{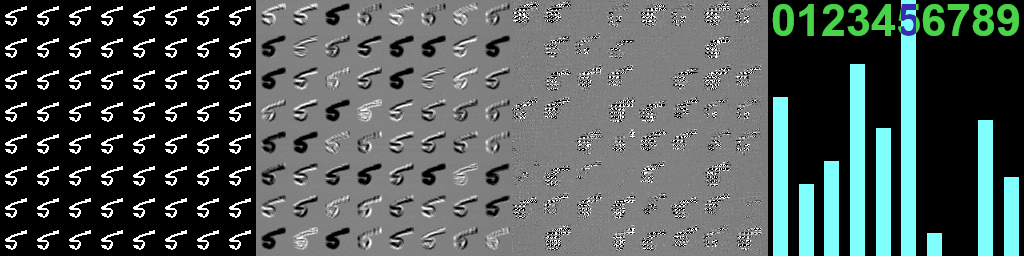} 
  \includegraphics[width=0.96\linewidth]{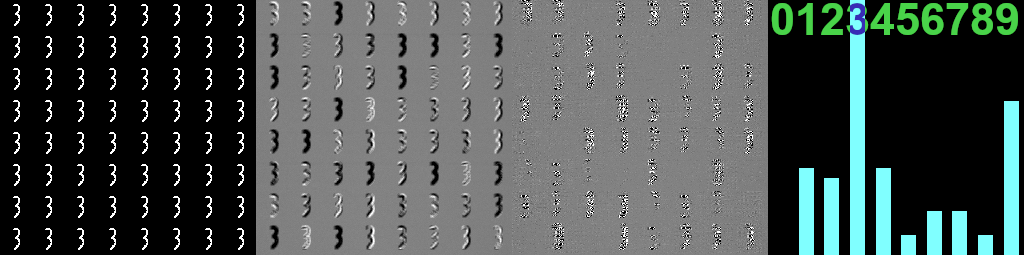} 

    \caption{Examples of digit classification via inference of a two layer network on SCAMP-5. showing both the 64 convolutions computed, parallel multiplication of fully connected weights and the activation values of the final fully connected layer's neurons, from 0 to 9. \label{fig:2layer classification examples}}
\end{figure*}

\subsection{Inference on SCAMP-5 Hardware}

We evaluated our inference approach using both two and three layer networks, trained on MNIST classification. The two layer networks used $32\times32$ input images, and consisted of one convolutional layer (64 feature maps, $4\times4$ filters), max pooling ($4\times4$), and a final fully connected layer. Three layer networks used up scaled $64\times64$ input images, a first convolutional layer (16 feature maps), max pooling ($4\times4$), a second convolutional layer (also 16 feature maps), and a final fully connected layer. Some sample classifications of such networks are shown in Figures \ref{fig:3layer classification examples} and \ref{fig:2layer classification examples}. Training is performed on a standard PC, using the 60,000 samples dataset. 
The trained weights were then loaded into the PEs of the SCAMP-5's processor array as described in previous sections, and  evaluation of inference was performed by directly loading testing set images (one at a time) onto the PE array.
With each image the SCAMP-5's then executed the SIMD routines to compute the network layers and output a final classification.
%Each loaded digit was copied to each computational block in the first convolutional layer as shown in Figure \ref{fig:convol examples}.
%The SIMD routine for convolutional layers was then performed once, followed by the routine for the max pooling, and the final fully connected layer, with the final ten global summation results producing the neuron activation values for each possible digit. 
Note that it was necessary to convert the MNIST testing images to 1-bit images when loading them directly into the PE array.
Table \ref{tab:timings} shows the computation times of the various processes used during inference. 

\begin{table}[h]
\centering
 \begin{tabular}{||c | c | c ||} 
 
%   \hline
%  \multicolumn{3}{|c|}{Network Computation Times  } \\
%  \hline
 
 \hline
 Component & Two Layer Network &  Three Layer Network \\ [0.5ex] 
 \hline\hline
Digit Duplication & 28$\mu s$  & 28$\mu s$ \\ 
 \hline
Convolutional Layer/s & 160$\mu s$ & 320$\mu s$ (160 $\times$ 2)\\
 \hline
ReLu & $<$1$\mu s$ & $<$1$\mu s$  \\
 \hline
 Max Pooling & 25$\mu s$ & 25$\mu s$\\
  \hline
Feature Map Shrink and Duplicate & - & 2095$\mu s$\\
 \hline
Feature Map Creation & - & 1055$\mu s$\\
 \hline
  Fully Connected Layer & 59$\mu s$ & 901$\mu s$\\
 \hline
  \textbf{Total} & 272$\mu s$ (3676 fps) & 4464$\mu s$ (224 fps)\\ [1ex] 
 \hline
\end{tabular}
\caption{Computation times of various network components.}\label{tab:timings}
\end{table}

The total computation time of two layer networks was 272 microseconds corresponding to the processing speed of 3676 classifications per second (excluding time to load the input to the array, but including the time to duplicate the input image across the array). It is assumed that in a real-time deployment scenario, images would not be loaded to the array, but rather obtained via the image sensing capabilities of the chip, with appropriate region-of-interest detection and cropping/rescaling, as demonstrated in \cite{bose2019camera}. The inference accuracy of tested two layer networks varied around 92\%-94\% classification accuracy with different networks, a reduction from accuracy levels around 95\% obtained in training on PC.
It is worth noting that due to the nature of the analog computing used during inference, which exhibits noise and systematic errors (\cite{carey2013100}) and which currently vary from one SCAMP-5 device to another, such a drop in accuracy and is not unexpected. 

Three layer networks had a total computation time of 4.46 miliseconds (giving 224 classifications per second), the majority of which was from the shrinking of activation data between convolution layers, and the merging of convolutions to form feature maps.
The methods and SIMD routines for these components are not as optimized as those for layer computation, and is something we seek to improve in the future.
The classification accuracy obtained for three layer network inference was also in the range of 92\%-94\%, but with a more significant drop from the accuracy of 97\% obtained in training.

It may be possible to reduce these discrepancies between training and inference accuracy in future work, either by modeling the analog errors within the training process, by using a harware-in-the-loop approach performing forward pass directly on SCAMP hardware during training, or by shifting certain components to use digital rather than analog computation upon the PPA.
That said, the PPAs massively parallel analog computing still results in high performance and efficiency. During inference vision sensor itself consumes 1.25 W, with the rest of the current camera system contributing another 750 mW, when operating at the maximum throughput of 3676 classifications per second. It can be extrapolated, that for applications where frame rates in the range of 30 fps are acceptable, the operating power of the system executing a two-layer network model would be in the range of 10-20 mW.

\newpage
\section{Conclusions}
We have presented a novel approach for conducting CNN inference upon PPA hardware, exploiting analog computations, and storing the weights of the network directly within the processing elements themselves rather than in the program running upon the processor array's controller chip.
Unlike previous works, our approach can perform multiple convolution layers, and a final fully connected layer entirely upon the PE array of the device. 
With the image sensing also carried out by the device, neither the images, nor their filtered versions, need to be ever transmitted off-chip. The only information read-out is the activations of the final neuron layer. Thus the system demonstrates a complete CNN on-chip solution, from light sensing, to classification results.  
Our experiments considered small network topologies using binary filters and ternary fully connected weights. Our approach can be applied to deeper more complex networks with additional convolutional layers and larger filter sizes, as PPA hardware improves. 

While our contribution is relevant beyond current PPAs, even smaller networks like the one we demonstrate here have found practical applications in the edge computing devices, and our approach demonstrates, for the first time, a complete classification task executed on the "focal-plane" of an image sensor device.
We demonstrate our approach via inference of digit classification networks, being performed at over 93\% classification accuracy. Our experimental camera system (including SCAMP-5 chip, and the associated control and interface circuits) operates at a speed enabling over 3,000 image classifications per second. It can be expected that implementing the hardware system in more recent silicon technologies will provide substantial gains in performance and efficiency.

Just as with their nature counterparts, fast, energy efficient sensor-processor arrays that are capable of learning to respond to what they sense are likely to play a significant role in taking visual competences into the demands of a world that gave rise to visual perception in the first place.

\newpage
% ---- Bibliography ----
%
% BibTeX users should specify bibliography style 'splncs04'.
% References will then be sorted and formatted in the correct style.
%
\bibliographystyle{splncs04}
\bibliography{main}
\end{document}